\documentclass[conference]{IEEEtran}
\usepackage{graphicx}
\usepackage{float}
\usepackage[caption=false]{subfig}
\usepackage[space]{grffile}
\usepackage{latexsym}
\usepackage{enumerate}
\usepackage{textcomp}
\usepackage{amsfonts,amsmath,amssymb}
\usepackage{bm}
\DeclareMathOperator*{\argmax}{\arg\!\max}
\usepackage{url}
\usepackage{algorithm}
\usepackage{algpseudocode}
\usepackage{tabularx}
\usepackage{longtable}
\usepackage{multirow,booktabs,makecell}
\usepackage{xcolor}

\newcommand{\BO}{GPBO}
\newcommand{\gp}{\mathcal{GP}(\fm,\fk)}
\newcommand{\fy}{y}

\newcommand{\K}[2]{K({#1},{#2})}
\newcommand{\Kx}{\K{\traindata}{\traindata}}
\newcommand{\Kxs}{\K{\traindata_*}{\traindata_*}}
\newcommand{\Kxxs}{\K{\traindata}{\traindata_*}}
\newcommand{\Kxsx}{\K{\traindata_*}{\traindata}}

\newcommand{\fv}{\bm{\mathrm{f}}}
\newcommand{\fs}{\mathcal{S}}
\newcommand{\fm}{m(\bovar)}
\newcommand{\fk}{k(\bovar,\bovar^\prime)}
\newcommand{\fa}{a(\bovar)}
\newcommand{\bovar}{\mathbf{q}}
\newcommand{\bovarspace}{\mathcal{Q}}
\newcommand{\traindata}{Q}
\newcommand{\Jnees}[0]{J_{NEES}}
\newcommand{\Jnis}[0]{J_{NIS}}

\def\mr[#1]#2#3{\multirowcell{#2}[#1]{#3}}

\newcommand{\kLst}{k-1}
\newcommand{\kCur}{k}

\newcommand{\x}[1]{\mathbf{x}_{#1}}
\newcommand{\z}[1]{\mathbf{z}_{#1}}
\newcommand{\uVec}[1]{\mathbf{u}_{#1}}
\newcommand{\vVec}[1]{\mathbf{v}_{#1}}
\newcommand{\wVec}[1]{\mathbf{w}_{#1}}
\newcommand{\F}[1]{\mathbf{F}_{#1}}
\newcommand{\Ft}[1]{\mathbf{F}_{#1}^\top}
\newcommand{\HM}[1]{\mathbf{H}_{#1}}
\newcommand{\HMt}[1]{\mathbf{H}_{#1}^\top}
\newcommand{\Q}[1]{\mathbf{Q}_{#1}}
\newcommand{\R}[1]{\mathbf{R}_{#1}}
\newcommand{\ex}[1]{\mathbf{e}_{\mathbf{x},#1}}
\newcommand{\ez}[1]{\mathbf{e}_{\mathbf{z},#1}}
\newcommand{\nees}[1]{\epsilon_{\mathbf{x},#1}}
\newcommand{\nis}[1]{\epsilon_{\mathbf{z},#1}}
\newcommand{\avgnees}[1]{\bar{\epsilon}_{\mathbf{x},#1}}
\newcommand{\avgnis}[1]{\bar{\epsilon}_{\mathbf{z},#1}}
\newcommand{\W}[0]{\mathbf{W}}
\newcommand{\V}[0]{\mathbf{V}} 
\newcommand{\A}[0]{\mathbf{A}}

\newcommand{\E}[1]{\mathrm{E}\left[#1\right]}
\newcommand{\EOuter}[1]{\E{{#1}{#1}^\top}}

\newcommand{\xCond}[2]{\hat{\mathbf{x}}_{#1|#2}}

\newcommand{\covCond}[2]{\mathbf{P}_{#1|#2}}

\newcommand{\zCond}[2]{\hat{\mathbf{z}}_{#1|#2}}

\newcommand{\innovCov}[2]{\mathbf{S}_{#1|#2}}
\newcommand{\nx}[0]{n_{\mathbf{x}}}
\newcommand{\nz}[0]{n_{\mathbf{z}}}
\newcommand{\Kw}[1]{\mathbf{K}_{#1}}
\newcommand{\Kwt}[1]{\mathbf{K}_{#1}^\top}

\newcommand{\Snu}[1]{\mathbf{S}_{#1}}

\begin{document}
\title{Weak in the NEES?: Auto-tuning Kalman Filters \\ with Bayesian Optimization}

\author{
\IEEEauthorblockN{Zhaozhong Chen, Christoffer Heckman}
\IEEEauthorblockA{Department of Computer Science \\ University of Colorado Boulder \\
430 UCB \\ Boulder, CO 80309 \\ email: Christoffer.Heckman@colorado.edu}
\and
\IEEEauthorblockN{Simon Julier}
\IEEEauthorblockA{Department of Computer Science \\ University College London\\
66--72 Gower Street, \\ London WC1E 6BT, UK \\ email: s.julier@ucl.ac.uk}
\and
\IEEEauthorblockN{Nisar Ahmed}
\IEEEauthorblockA{Smead Aerospace Engineering Sciences \\ University of Colorado Boulder \\
429 UCB \\ Boulder, CO 80309 \\ email: Nisar.Ahmed@colorado.edu}
}
\maketitle

\begin{abstract}
Kalman filters are routinely used for many data fusion applications including
  navigation, tracking, and simultaneous localization and mapping problems.
  However, significant time and effort is frequently required to tune various
  Kalman filter model parameters, e.g.\ process noise covariance, pre-whitening
  filter models for non-white noise, etc.
Conventional optimization techniques for tuning can get stuck in poor local
minima and can be expensive to implement with real sensor data. To address
these issues, a new ``black box'' Bayesian optimization strategy is developed for
automatically tuning Kalman filters. In this approach, performance is
characterized by one of two stochastic objective functions: normalized
estimation error squared (NEES) when ground truth state models are available,
or the normalized innovation error squared (NIS) when only sensor data is
available. By intelligently sampling the parameter space to both learn and
exploit a nonparametric Gaussian process surrogate function for the NEES/NIS
costs, Bayesian optimization can efficiently identify multiple local minima and
provide uncertainty quantification on its results.
\end{abstract}

\IEEEpeerreviewmaketitle

\section{Introduction}

Although research in the past few years has introduced many new estimation
algorithms, the Kalman filter still remains one of the most widely used
algorithms in the world today. Its popularity can largely be attributed to its
efficiency, simplicity and robustness.

A major challenge in developing a Kalman filter is that it must be
\emph{tuned\/}. Given a real world application and a system design, the process
and observation covariance matrices must be set to given an acceptable level of
performance. This performance is often defined in terms of the mean squared
error of the estimate. The general idea behind tuning is to search over the
space of filter parameters and assess performance. If one has access to ground
truth (either from an extra measurement system or simulation), the performance
can be assessed statistically based on the normalized estimation error squared
(NEES). However, often only observation sequences are available and thus the
normalized innovation squared  (NIS) must be used instead. Tuning then becomes
a problem of balancing the behaviour of the filter performance metric over
time. One approach is to do this manually, i.e. simply explore over all
available degrees of freedom until good results obtained. However, this can
often be a long and difficult process which requires studying the interaction
of many different filter parameters. 

Given the difficulty of manual tuning, methods for automating filter tuning are
of great practical interest. These methods typically pose tuning as an
optimisation problem: given a measure of performance, such as NIS and NEES,
iterate through points in parameter space to find the one which provides the
best results. These can give very good results. However, a key issue is that
the optimisation problem is often highly non-covex. As a result, gradient-based
optimization algorithms suffer from the possibility that they could fall into a
local minima.

In this paper, we consider the problem of how to develop Kalman filter tuning
algorithm using Bayesian Optimization. Our idea is to recast optimization as a
Bayesian search problem in which the next iteration of the optimizer seeks a point
which maximizes the probability of improving an overall measure of the state
estimator performance. As such, Bayesian optimization offers a potentially
principled way to handle the local minima problem. For this initial
investigation, we restrict ourselves to linear systems. However, the underlying
principles apply to nonlinear systems as well and are amenable to extension
covering these cases.

This paper is structured as follows. Section~\ref{sct:problem_statement}
introduces the filter tuning problem.  Section~\ref{sct:bayesopt} provides an
overview of Bayesian optimization using Gaussian process models for optimizing
stochastic black box cost functions, and then describes its novel application
to Kalman filter tuning using cost functions based on $\chi^2$ consistency test
statistics. 
Section~\ref{sct:results} presents numerical examples showing the application
of Bayesian optimization auto-tuning to a linear system.  Conclusions and
ongoing/ future work are given in Section~\ref{sct:conc}.

\section{Preliminaries}
\label{sct:problem_statement}

\subsection{System Description}

Consider the problem of estimating the state and quantifying the uncertainty in that estimate in discrete time. Let the state of the system at time step $k$ be $\x{k}$; our goal is to develop an  algorithm that can result in the state estimate. Let $\xCond{i}{j}$ be the estimate of $\x{i}$ using all observations up to time step $j$, and the covariance of this estimate be $\covCond{i}{j}$:
\begin{align}
\xCond{i}{j}&=\E{\x{i}|\z{1:j}}\\
\covCond{i}{j}&=
\EOuter{\left(\x{i}-\xCond{i}{j}\right)|\z{1:j}}.
\end{align}

The system is described by a process model and an observation model. The  process model that describes how the system evolves from state $\kLst$ to $\kCur$ is:
\begin{equation}
\x{\kCur}=\F{\kCur}\x{\kLst}+\mathbf{B}_{\kCur}\uVec{\kCur}+\vVec{\kCur},
\label{eq:dynModel}
\end{equation}
\noindent where $\uVec{\kCur}$ is the control input and $\vVec{\kCur}$ is the process noise, which is assumed to be zero
mean and independent with covariance $\Q{\kCur}$. The observation model is
\begin{equation}
\z{\kCur}=\HM{\kCur}\x{\kCur}+\wVec{\kCur}, \label{eq:measModel}
\end{equation}
\noindent where $\wVec{\kCur}$ is the observation noise. This is assumed to be zero-mean and independent with a covariance $\R{\kCur}$.

As is well-known, a Kalman filter may be applied to this problem in order to find the optimal estimate \cite{Kalman-JBE-1961}; this filter follows a two stage process of prediction followed by update. The predicted state is given by
\begin{align}
\xCond{\kCur}{\kLst}&=\F{\kCur}\xCond{\kLst}{\kLst} +\mathbf{B}_{\kCur}\uVec{\kCur}\\
\covCond{\kCur}{\kLst}&=\F{\kCur}\covCond{\kLst}{\kLst}\Ft{\kCur}+\Q{\kCur}
\end{align}
\noindent while the update is given by 
\begin{align}
\xCond{\kCur}{\kCur}&=\xCond{\kCur}{\kCur}+\Kw{\kCur}\ez{\kCur},\\
\covCond{\kCur}{\kCur}&=\covCond{\kCur}{\kLst}-\Kw{\kCur}\Snu{\kCur|\kLst}\Kwt{\kCur},\\
\Snu{\kCur|\kLst}&=\HM{\kCur}\covCond{\kCur}{\kLst}\HMt{\kCur}+\R{\kCur}\\
\Kw{\kCur}&=\covCond{\kCur}{\kLst}\HMt{\kCur}\Snu{\kCur|\kLst}^{-1}
\end{align}
where $\ez{\kCur}=\zCond{\kCur}{\kLst} - \z{\kCur}$ is the so-called \emph{innovation vector\/}.

A dynamical state estimator is statistically consistent if the following conditions are met \cite{Bar-Shalom2001}: 
\begin{enumerate}
\item the state estimation errors are unbiased,
\begin{align}
\E{\ex{k}} = \mathbf{0}, \ \forall k
\end{align}
\item the estimator is efficient, 
\begin{align}
\E{\ex{k}\ex{k}^T} = \covCond{k}{k}, \ \forall k
\end{align}
\item the innovations form a white Gaussian sequence, such that for all times $k$ and $j$,
\begin{align}
\ez{k} &\sim {\cal N}(\mathbf{0},\innovCov{k}{k-1})\\
\E{\ez{k}} &= \mathbf{0}, \\
\E{\ez{k}\ez{j}^T} &= \delta_{jk}\cdot \innovCov{k}{k-1}
\end{align}
\end{enumerate}

Intuitively, a filter is statistically consistent if it correctly describes the actual state error statistics for any set of (simulated) ground truth state sequences, as well as correctly describes the actual measurement residual errors for any set of measurement data logs. When the full structure of the system state ($\F{\kCur}$, $\HM{\kCur}$,  $\Q{\kCur}$ and $\R{\kCur}$) is known, the Kalman filter equations automatically guarantee statistical consistency. However, in many situations the model is not known precisely, and so the filter must be tuned.

\subsection{Filter Tuning}

Filter tuning is the process of selecting parameters to optimize performance. Consistency ensures two desirable properties in a Kalman filter: (i) the filter is `aware' of how wrong it could actually be; and (ii) the filter blends the right amount of information from its process model and measurements to recursively correct its state estimate. 

Given values for $\F{\kCur}$ and $\HM{\kCur}$, tuning involves choosing $\Q{\kCur}$ and $\R{\kCur}$. If the model is matched ($\F{\kCur}$ and $\HM{\kCur}$ are the same as the true system), the statistical consistency can be achieved. However, in general the model can be mismatched. In this case, we seek to satisfy the weaker condition of \emph{covariance consistency},
\begin{align}
\xCond{i}{j}&\approx\E{\x{i}|\z{1:j}}\\
\covCond{i}{j}&\ge
\EOuter{\left(\x{i}-\xCond{i}{j}\right)|\z{1:j}}.
\end{align}
where $\approx$ is application specific and $\mathbf{A}\ge\mathbf{B}$ means that $\mathbf{A}-\mathbf{B}$ is positive semidefinite. In other words, the estimate should be approximately unbiased, and the estimator should not over estimate its level of confidence. At the same time, the estimated covariance should not be very large.

These conditions can be assessed by examining the normalized scalar magnitudes of the random variables $\ex{k}$ and $\ez{k}$, 
\begin{align}
\nees{k} &= \ex{k}^T \covCond{k}{k}^{-1} \ex{k} \label{eq:neesDef}  \\
\nis{k} &= \ez{k}^T \innovCov{k}{k-1}^{-1} \ez{k} \label{eq:nisDef},
\end{align}
which define the \emph{normalized estimation error squared (NEES)} and \emph{normalized innovation error squared (NIS)}, respectively. If the dynamical consistency conditions are met, then it is easy to show that $\nees{k}$ and $\nis{k}$ should be $\chi^2$ random variables with $\nx$ and $\nz$ degrees of freedom, respectively \cite{Bar-Shalom2001}. 
Therefore, $\chi^2$ hypothesis tests can be performed on calculated values for $\nees{k}$ (when ground truth data is available) and $\nis{k}$ to see if the consistency conditions hold at each time $k$. 

In practice, NEES $\chi^2$ tests are conducted using multiple offline Monte Carlo `truth model' simulations to obtain ground truth $\x{k}$ values. The truth model simulator represents a high-fidelity model of the `actual' system dynamics and sensor observations, which may contain non-linearities and other non-ideal characteristics that must be compensated for via Kalman filter tuning. NIS $\chi^2$ can be conducted offline using multiple Monte Carlo simulations (e.g. in parallel with NEES tests), but can also be conducted online with real sensor data logs. 

Offline truth model tests are conducted as follows \footnote{Online NIS tests with real sensor data are similar, but exploit ergodicity of measurement innovation sequences}: suppose $N$ independent instances of the true state are randomly initialized according to $\xCond{0}{0}$ and $\covCond{0}{0}$ (the initial state of the filter), and then propagated through the true stochastic dynamics \eqref{eq:dynModel} and measurement model \eqref{eq:measModel} for $T$ time steps, yielding sample ground truth sequences $\x{1}^i,\x{2}^i,\ldots,\x{T}^i$ and measurement sequences $\z{1}^i,\z{2}^i,\ldots,\z{T}^i$ for $i=1,\ldots,N$. 
If the resulting measurement sequences are then fed to a Kalman filter with tuning parameters $(\Q{k},\R{k})$,  the resulting NEES and NIS statistics for each simulation run $i$ at each time $k$ can be averaged across problem instances to give the test statistics
\begin{align}
\avgnees{k} &= \frac{1}{N} \sum_{i=1}^{N}{\nees{k}^i} \label{eq:neesAvgk} \\
\avgnis{k} &= \frac{1}{N} \sum_{i=1}^{N}{\nis{k}^i}. \label{eq:nisAvgk}
\end{align}
Then, given some desired Type I error rate $\alpha$, the NEES and NIS $\chi^2$ tests provide lower and upper tail bounds $[l_{\mathbf{x}}(\alpha,N),u_{\mathbf{x}}(\alpha,N)]$ and $[l_{\mathbf{z}}(\alpha,N),u_{\mathbf{z}}(\alpha,N)]$, such that the Kalman filter tuning is declared to be consistent if, with probability $100(1-\alpha)$ at each time $k$,
\begin{align*}
\avgnees{k} \in [l_{\mathbf{x}}(\alpha,N),u_{\mathbf{x}}(\alpha,N)] 
\mbox{ \ and \ }  
\avgnis{k} \in [l_{\mathbf{z}}(\alpha,N),u_{\mathbf{z}}(\alpha,N)].
\end{align*}
Otherwise, the filter is declared to be inconsistent.  Specifically, if $\avgnees{k}<l_{\mathbf{x}}(\alpha,N)$ or $\avgnis{k}<l_{\mathbf{z}}(\alpha,N)$, then the filter tuning is `pessimistic' (`underconfident'), since the filter-estimated state error/innovation covariances are too large relative to the true values.  
On the other hand, if $\avgnees{k}>u_{\mathbf{x}}(\alpha,N)$ or $\avgnis{k}>u_{\mathbf{z}}(\alpha,N)$, then the filter tuning is `optimistic' (`overconfident'), since the filter-estimated state error/innovation covariance are too small relative to the true values. 

The $\chi^2$ consistency tests provide a very principled basis for validating Kalman filter performance in domain-agnostic way, and also provide a well-established means for guiding the tuning of noise parameters $\Q{k}$ and $\R{k}$ in practical applications. Tuning via the $\chi^2$ tests is most often done manually, and thus requires repeated `guessing and checking' over multiple Monte Carlo simulation runs. However, this quickly becomes cumbersome and non-trivial for systems with several tunable noise terms. Heuristics for manual filter tuning have been developed in the linear-quadratic optimal control literature \cite{stengel1986optimal}, e.g. to coarsely tune diagonals of $\Q{k}$ first, before fine-tuning the elements of $\Q{k}$ further. Such heuristics are useful for bounding the shape and magnitude of $\Q{k}$ in linear-Gaussian problems, but are of little help for tuning `fudge factor' process noise parameters that are used to cope with model errors from state truncation, approximations of non-linearities, poorly modeled dynamics, etc. 

Manual tuning is especially challenging if truth model simulations are computationally expensive to run or the filter involves many parameters which can interact with one another in subtle and surprising ways. This not only also makes it difficult to explore the parameter space to properly calibrate heuristics, but also makes it difficult to achieve a large enough $N$ to properly assess inherently noisy NEES/NIS test statistics. Furthermore, because the NEES and NIS are outputs of a stochastic non-differentiable `black box' simulation function, the filter tuning process cannot be simply automated via conventional convex optimization methods (e.g. line search, gradient descent, etc.). Given this, we need to use alternative optimization techniques which are robust to stochastic variations in the cost function and can explore nonlinear spaces.

\section{Bayesian Optimization for Filter Auto-tuning} \label{sct:bayesopt}

A common approach to solving nonlinear optimization problems is to use gradient descent. However, the risk with these approaches is that they can fall into local minima. This issue is exacerbated for filter tuning problems where objective functions are governed by noisy dynamical systems. With a finite number of samples, stochastic variation introduces many small local minima and maxima which can trap gradient descent methods. One principled way to handle parameter tuning problems in such cases is to use \emph{Bayesian Optimization\/} which poses optimization as a Bayesian search problem. The objective function is unknown and is treated as a random variable. A prior is placed over it. As the algorithm proceeds, each iteration takes samples from the objective function which are used to refine the distribution. The next sample point is selected to maximize the probability of improving the current best estimate.

First we will describe Bayesian optimization for dealing with generic `black box' stochastic objective functions. We then describe its novel application for simulation-based Kalman filter auto-tuning. 

\subsection{Bayesian Optimization Theory}

Consider the minimization of some objective function $y:\bovarspace \rightarrow \mathbb{R}$, where $\bovarspace \in \mathbb{R}^d$ is the search or solution space, and the element $\bovar^* \in \bovarspace$ is the minimizer, such that $y(\bovarspace^*) \leq y(\bovarspace), \  \forall \bovar \in \bovarspace$.
For simplicity, we assume the solution space is bounded for global optimization, where $\bovarspace (i) \in [\bovar(i)_l, \bovar(i)_u]$ for lower bound $\bovar(i)_l$ and upper bound $\bovar(i)_u$ for element $i$ of $\bovar$. 
When the mapping from $\bovar$ to $\fy$ is not known explicitly, the optimization typically requires the evaluation of a `black box' function. 
In our application, $\fy$ is the result of evaluating the performance of a Kalman filter in design configuration $\bovar$ on a set of synthetic/real sensor data logs generated by a `true' underlying dynamical system. The black box evaluations of $\fy$ can therefore be expensive, slow, and produce noisy results for the same input $\bovar$.

The goal of Bayesian optimization is to find the minimizer of a noisy objective function $\fy$ that is costly to evaluate at any given design point $\bovar$, while also learning about the mapping from $\bovar$ to $\fy$ at the same time via Bayesian inference. An initial prior belief $p(y)$ over possible $\fy$ functions is updated by subsequent observations (evidence) $E$ consisting of sample $\fy$ evaluations for different sampled $\bovar$ values. Mathematically, this leads to an application of Bayes' rule: $p(y|E) \propto p(E|y)p(y)$, where $p(E|y)$ is the observation likelihood and $p(y|E)$ is the posterior of $\fy$ given $E$. Hence, evidence $E$ gives information about the actual shape of $\fy$, allowing the posterior belief about the assumed shape of $\fy$ to be recursively updated. As long as both $p(y)$ and $p(y|E)$ are consistent with the true nature of $\fy$, then the law of large numbers ensures that the posterior $p(y|E)$ converges with high probability to the true $\fy$, in the limit of infinite observations $E$ covering $\bovarspace$. 

Bayesian optimization uses black box point evaluations of $y$ to efficiently find $\bovar^*$. This is accomplished by maintaining beliefs about how $\fy$ behaves over all $\bovar$ in the form of a ``surrogate model'' $\fs$, which statistically approximates $\fy$ and is easier to evaluate (e.g. since $\fy$ might be an expensive high fidelity simulation). During optimization, $\fs$ is used to determine where the next design point sample evaluation of $\fy$ should occur, in order to update beliefs over $\fy$ and thus simultaneously improve $\fs$ while finding the (expected) minimum of $\fy$ as quickly as possible. The key idea is that, as more observations are sampled at different $\bovar$ locations, the $\bovar$ samples themselves eventually converge to the expected minimizer $\bovar^*$ of $\fy$. Since $\fs$ contains statistical information about the level of uncertainty in $\fy$ (i.e. related to $p(y|E)$, the posterior belief), Bayesian optimization effectively leverages probabilistic `explore-exploit' behavior to learn an approximate model of $\fy$ while also minimizing it. We next describe the two main components of the Bayesian optimization process: (1) the surrogate model $\fs$, which encodes statistical beliefs about $\fy$ in light of previous observations and a prior belief; and (2) the acquisition function $\fa$, which is used to intelligently guide the search for $\bovar^*$ via $\fs$. 

\subsubsection{The Surrogate Model}\label{sec:gaussian-processes}
$\fs$ must approximate $\fy$ in areas where it has not yet been evaluated, and must also provide a predicted value and corresponding uncertainty to quantify the possibility that the optimum is located at some location $\bovar$.  
Gaussian Processes (GPs)~\cite{Rasmussen2006} are the most common family of surrogate models used in Bayesian optimization; 
 the acronym \BO{} here refers to Bayesian optimization using a GP surrogate model $\fs$. A GP describes a distribution over functions; it is more formally defined as a collection of random variables, any finite number of which have a joint Gaussian distribution ~\cite{Bishop2006,Rasmussen2006},  
    \begin{align}
        f(\bovar) &\sim \gp \label{eq:gp_basic} \\
        \fm &= \mathbb{E}[f(\bovar)] \label{eq:gp_mean} \\
        \fk &= \mathbb{E}[(f(\bovar)-m(\bovar))(f(\bovar^\prime)-m(\bovar^\prime))] \label{eq:gp_cov}
    \end{align}
where the process is completely specified by its mean function $\fm$ (equation~\ref{eq:gp_mean}), and its covariance function $\fk$ (equation~\ref{eq:gp_cov}).     
In theory $\fm$ could be any function; as is common practice, this work assumes $m$ is zero for simplicity. The covariance (or kernel) function is a mapping $k:(\bovar,\bovar^\prime) \rightarrow \mathbb{R}$; this must be specified a priori, and is usually based on some knowledge of $\fy$'s smoothness properties.

A valid kernel must be positive semi-definite (PSD), i.e. it must produce a Gram matrix $K$, with individual elements $[K_{i,j}]$ given by $k(\bovar_{i},\bovar_{j})$, that is PSD given a set of training data $\traindata = \{\bovar_{1},\ldots,\bovar_n\}$. Let $\Kx$ be the Gram matrix defined by kernel function $k$, 
    \begin{align}
        \Kx{} &= \begin{bmatrix}
                k({\bovar}_1,{\bovar}_1) & k({\bovar}_1,{\bovar}_2) & \cdots & k({\bovar}_1,{\bovar}_n)  \\
								k({\bovar}_2,{\bovar}_1) & k({\bovar}_2,{\bovar}_2) & \cdots & k({\bovar}_2,{\bovar}_n)  \\
								\vdots		 &  \vdots    &   \vdots    \\ 
                k({\bovar}_n,{\bovar}_1) & k({\bovar}_n,{\bovar}_2) & \cdots & k({\bovar}_n,{\bovar}_n)
					   \end{bmatrix}.
    \end{align}
Given $n$ training observations, the elements of the covariance matrix $\Kx \in \mathbb{R}^{n\times n}$ are the covariances $k({\bovar}_i,{\bovar}_j)$ between ${\bovar}_i$ and ${\bovar}_j$ for all pairs of training data. 
The joint distribution of $n$ training outputs $\fv(\traindata)\in\mathbb{R}^{n\times 1}$ and $p$ test outputs $\fv_{*}(\traindata_*)\in\mathbb{R}^{p\times 1}$ for inputs $ \traindata_*=\{\bovar_{*1},\ldots,\bovar_{*p}\}$ is 
	\begin{align}
        \begin{bmatrix}
            \fv \\
            \fv_* \end{bmatrix} &\sim  \mathcal{N}\left(\textbf{0},\begin{bmatrix}
                \Kx & \Kxxs \\
                \Kxsx & \Kxs \end{bmatrix}\right) \label{eq:fX}, \\
						\Kxsx & = \begin{bmatrix} 
						k(\bovar_{*1},\bovar_1) & k(\bovar_{*1},\bovar_2) & \cdots & k(\bovar_{*1},\bovar_n) \\
						k(\bovar_{*2},\bovar_1) & k(\bovar_{*2},\bovar_2) & \cdots & k(\bovar_{*2},\bovar_n) \\
						\vdots & \vdots & \ddots & \vdots \\
						k(\bovar_{*p},\bovar_1) & k(\bovar_{*p},\bovar_2) & \cdots & k(\bovar_{*p},\bovar_n) \end{bmatrix}
    \end{align}
Given $\traindata$ and $\fv$, $\fv_*$ can be predicted at new `test locations' $\bovarspace_*$, using the conditional GP mean and covariance relations 
    \begin{align}
        &\fv_{*}|\traindata_{*},\traindata,\fv \sim \mathcal{N}(\mu(\traindata_*),\sigma^2(\traindata_*)) \label{eq:predict2} \\
        &\mu(\traindata_*) = \Kxsx \Kx^{- 1}\fv \label{eq:mu}\\
        &\sigma^2(\traindata_*) = \Kxs - \Kxsx \Kx^{- 1} \Kxxs \label{eq:sigma}
    \end{align}
Here, $\Kxxs \in \mathbb{R}^{p \times n}$, so that $\mu(\traindata_*)\in\mathbb{R}^{p\times 1}$ and $\sigma^2(\traindata_*)\in\mathbb{R}^{p \times p}$. Eq.\ \eqref{eq:predict2} gives the expression of the conditional distribution of $\fv_*$ given test points $\traindata_*$, and training data $\traindata$ and $\fv$. The mean and variance of this predictive distribution are found via Eqs.\ \eqref{eq:mu} and \eqref{eq:sigma}. In the context of Bayesian optimization, the GP surrogate model provides statistical information (i.e. mean and variance from \ref{eq:mu} and \ref{eq:sigma}) of how the underlying objective function $\fy$ behaves for all possible values $\traindata_*$ that have not yet been sampled. 

The Mate\'rn kernel is one of the most popular choices for the kernel function $k$ in \BO, 
\begin{align}
k_{\nu = 3/2}\left( \bm{x}_{b,i}, \bm{x}_{b,j} \right) &= \sigma_0\left( 1 + \frac{\sqrt{3} r_{ij}}{\ell} \right)\exp\left( - \frac{\sqrt{3}r_{ij}}{\ell} \right), \\
\label{eq:Mat32}
r_{ij} &= \sqrt{(\bm{x}_{b,i} - \bm{x}_{b,j})^T(\bm{x}_{b,i} - \bm{x}_{b,j})},
\end{align}
with hyperparameters $\sigma_0$ and $\ell $, which are the kernel amplitude and length-scale, respectively. 
This kernel is guaranteed to be $k$ times differentiable when $k\leq \nu$ (where $\nu$ is nearly always taken to be half integer to simplify the kernel expression). 
As is standard in GP regression, an additive observation noise variance $\sigma_n^2$ is also assumed for each training datum $f(\bovar_{i})$, where $\bovar_{i} \in \traindata$
\begin{align}
f(\bovar_{i}) &= y(\bovar_{i}) + \epsilon_i, \label{eq:noisy_var}\\ 
\epsilon_i &\sim {\cal N}(0, \sigma^2_n), \label{eq:noise}
\end{align}
Hence, the full set of hyperparameters $\Theta = \left\{\sigma_n^2, \sigma_0, \ell \right\}$ governs the GP covariance function in Eq.\ \eqref{eq:gp_cov}. 

Since the best $\Theta$ setting is not known a priori, it must be learned and updated during \BO{}. Point estimation strategies based on maximum likelihood estimation and maximum a posteriori estimation are the most widely used in the \BO{} literature for supervised learning of $\Theta$ ~\cite{Shahriari2015a}. 
Fast gradient-based convex optimization techniques are most commonly used to minimize the negative log likelihood, since the required derivatives can be obtained analytically. 
However, since the GP likelihood is generally non-convex, numerical optimization can converge to many different local optima for $\Theta$. 
Furthermore, the best local optimum may be undesirable for learning with sparse data early on in the \BO{} process, since the associated $\Theta$ values typically overfit the training data~\cite{Cawley2007, Rasmussen2006}. 
This behavior is especially important to consider when trying to minimize the number of simulations for \BO{} \cite{Israelsen-JAIS-2018}.

\subsubsection{The Acquisition Function}\label{acq_fxn}
The acquisition function is defined as the mapping $a:(\bovar, \fs) \rightarrow \mathbb{R}$, abbreviated as 
\begin{align}
        \fa &\triangleq a(\bovar,\gp) \label{eq:acq_bo}
    \end{align}
which assumes the inclusion of the GP surrogate model as an argument. 
\BO{} selects $\hat{\bovar} = \argmax_{\bovarspace}\fa$ as the next location in $\bovarspace$ to be evaluated in the search process. 
Ideally, $\fa$ should enable exploration and modeling of $\fy$ by sampling new locations $\bovar$ that will improve the accuracy of $\fs$. At the same time, $\fa$ must exploit $\fs$ to reach the expected minimum of $\fy$ as quickly as possible. Therefore, $\fa$ should not lead to greedy or myopic behavior, or get stuck in poor local minima. There are many ways to define $\fa$ to balance these needs, but the best choice is heavily application dependent ~\cite{Shahriari2015a,Brochu2010,Hoffman2011}. 
Some popular methods include Expected Improvement (EI) and the Upper Confidence Bound. We focus only on EI here, since it does not require extra hyperparameters. 

EI  selects the next sample point to maximize the statistically expected improvement in the optimum when when the current best minimizer is $\bovar^+$. The EI function is defined by ~\cite{Jones1998} 
    \begin{align*}
        a(\bovar) &= \begin{cases}
                            (\mu(\bovar) - \fv(\bovar^{+}))\Phi(Z) + \sigma(\bovar)\phi(Z) &,  \quad \sigma(\bovar)>0 \\
                            0 &, \quad \sigma(\bovar)=0
                        \end{cases} \label{eq:EI1}        \\
        Z &= \frac{\mu\left( \bovar \right) - \fv(\bovar^{+})}{\sigma(\bovar)}, 
    \end{align*}
where $\mu(\bovar)$ is the mean predicted value of the GP at $\bovar$ and $\sigma(\bovar)$ is the predicted standard deviation at $\bovar$, $\fv(\bovar^{+})$ is the best observed value of the objective function, 
and $\Phi(Z)$ and $\phi(Z)$ are the PDF and CDF of the standard normal distribution $Z$. 

For any definition of $\fa$, another optimization routine must be used to identify the maximum of $\fa$ via point-based evaluation on $\fs$. The most popular method for doing this in \BO{} is the DIviding RECTangles (DIRECT) algorithm \cite{Jones1998}, which is a fast global non-convex optimization method that uses the Lipschitz continuity properties of $\fs$ to bound function values in local rectangles and search accordingly for the best local maximum of $\fa$. Note that the use of a non-convex optimization technique like DIRECT makes sense here, since they key idea behind Bayesian optimization is that evaluation of $\fa$ at multiple test points $\bovar$ will be cheaper and faster than evaluating $\fy$ at those points directly. 
In this work, we use the classical approach of selecting a single new design point $\bovar$ on each iteration of \BO{}, although variations to sample multiple design points at once or repeatedly on each iteration are also possible ~\cite{Israelsen-JAIS-2018}.  

\subsection{Stochastic Costs for Consistency-based Filter Auto-tuning}
We now consider how $\fy(\bovar)$ can be defined via NEES and NIS consistency test statistics for Kalman filter tuning. As such, let $\bovarspace$ be some space of configurable Kalman filter parameters (e.g. the set of all parameters defining some positive definite symmetric process noise covariance $\Q{k}$) and let $\bovar \in \bovarspace$ be a design point. 

Consider first the case of tuning based on assessment of NEES statistics obtained via Monte Carlo ground truth simulation models. If $N$ Monte Carlo simulations are performed for $T$ time steps at any given design point $\bovar$, starting from the initial conditions $\xCond{0}{0}$ and $\covCond{0}{0}$, then the average NEES statistic $\avgnees{k}$ can be computed via (\ref{eq:neesAvgk}) for each time $k=1,...,T$. 
To summarize how `well-behaved' $\avgnees{k}$ is across all time steps, we can leverage the fact that the expected value of $\avgnees{k}$ for a consistent Kalman filter ought to be $\nx$, i.e. the degrees of freedom of the $\chi^2$ NEES random variable (which is the same as the number of states). We can therefore use the following scalar function $\fy(\bovar)$ to assess how much $\avgnees{k}$ deviates from this ideal expected value across all time steps $k$ in $N$ Monte Carlo truth model simulations evaluated at $\bovar$, 
\begin{align}
\fy(\bovar) = \Jnees(\bovar) =  \sqrt{ \left[ \log \left(\frac{\sum_{k=1}^{T}{\avgnees{k}}}{\nx} \right) \right]^2} 
\end{align}

By similar reasoning, we can also define      
\begin{align}
\fy(\bovar) = \Jnis(\bovar) =  \sqrt{ \left[ \log \left(\frac{\sum_{k=1}^{T}{\avgnis{k}}}{\nz} \right) \right]^2}.
\end{align}
where $\avgnis{k}$ could either represent NIS outcomes obtained from truth model simulation or from a set of real data logs. 

Many other possible cost functions could also be used to summarize the behavior of the NEES/NIS statistics relative to $\nx$. For instance, instead of the mean over $T$ steps, $\fy(\bovar)$ could be defined in terms of the min/max or median of $\avgnees{k}$ or $\avgnis{k}$ vs. $\nx$ over $T$ steps. 
Or, $\fy(\bovar)$ could also be based on counting the number of times $\avgnees{k}$ or $\avgnis{k}$ exceed the $\chi^2$ hypothesis test bounds $[l(\alpha,N),u(\alpha,N)]$ for some given $\alpha$. 
While such alternative cost definitions could be useful for different applications (say, depending on the filter parameters being tuned), we focus on $\Jnees$ and $\Jnis$ here for simplicity. 

Algorithm \ref{alg:BayesOpt} summarizes the \BO{} procedure for Kalman filter tuning. The termination criteria could be based on iteration thresholds, tolerances on changes to the optimum $\bovar$ and/or $\fy$ between iterations, or other methods. 
An attractive feature of \BO{} is that eqs. \ref{eq:gp_mean}-\ref{eq:gp_cov} naturally provide uncertainty quantification on the shape of the objective function at both sampled and unsampled locations. This allows \BO{} to cope with multiple local minima in the parameter space $\bovarspace$. 
However, in practice, the \BO's performance depends on the selection and parameterization of the surrogate model kernel, as well as the number and placement of initial training observations (i.e. seed points) to bootstrap the search process. 

{\centering
\begin{minipage}{.92\linewidth}

 \begin{algorithm}[H]
            \caption{\BO{} for Kalman Filter tuning}
            \label{alg:BayesOpt}
            \begin{algorithmic}[1] 
                    \State Initialize GP with seed data $\left\{\bovar_s, \fy_s\right\}_{s=1}^{N_{seed}}$ and hyperparameters $\Theta$
                    \While {termination criteria not met}
                        \State $\bovar_j = \argmax_{\bovarspace} \fa$ 
                        \State Evaluate $y(\bovar_j)$, e.g. using $\Jnees(\bovar)$ or $\Jnis(\bovar)$. 
                        \State Add $y(\bovar_j)$ to $\fv(Q)$, $\bovar_j$ to $Q$, and update $\Theta$ \label{ln:testing}
                    \EndWhile \\
                    \Return $\bovar^* = \arg \min_{\bovar_j \in Q} \fv(\bovar_j)$
            \end{algorithmic}
        \end{algorithm}
\end{minipage}
\par
}

\section{Numerical Application Examples} \label{sct:results}
For ease of
presenting the proof of concept and discussion in this initial investigation, we
restrict ourselves to an application case study involving a simple linear
time-invariant system. However, the underlying principles apply to more complex
linear and nonlinear systems as well.

Consider a robot that moves along a 1D track and receives position measurements
every $\Delta t = 0.1 s$. Suppose the position and velocity state $\x{} = [\xi,
\dot{\xi}]^T$ are governed by the linear time invariant kinematics model
\begin{align*}
\dot{\mathbf{x}}_t &= \mathbf{A} \x{t} + \mathbf{G} \uVec{t} + \mathbf{\Gamma} \vVec{t} \\
\z{t} &= \HM{} \x{t} + \wVec{t},
\end{align*}
\noindent where 
\begin{align*}
\mathbf{A} =
\begin{bmatrix}
0 & 1 \\
0 & 0 
\end{bmatrix}, \ \ 
\mathbf{G} =
\begin{bmatrix}
0 \\
1  
\end{bmatrix}, \ 
\HM{} =
\begin{bmatrix}
1 & 0  
\end{bmatrix}, \ 
\mathbf{\Gamma} = 
\begin{bmatrix}
0 \\ 
1  
\end{bmatrix}, \ 
\end{align*}
\noindent and the inputs to the system consist of a control acceleration $\uVec{t}$, additive white Gaussian noise acceleration process $\vVec{t}$ with intensity $\V$, and additive white Gaussian position measurement noise process $\wVec{t}$ with continuous time intensity $\W$. The control input $\uVec{t}=2\cos(0.75t)$ causes the robot to move with a low frequency oscillation. Applying a zero-order hold discretization to this system, we obtain discrete time position and velocity state $\x{\kCur} = [\xi_{\kCur}, \dot{\xi}_{\kCur}]^T$ and the linear time-invariant parameters for eqs. (\ref{eq:dynModel})-(\ref{eq:measModel})  
\begin{align*}
\F{} = 
\begin{bmatrix}
1 & \Delta t \\
0 & 1
\end{bmatrix}, \ \
\mathbf{B} = 
\begin{bmatrix}
0.5 \Delta t^2 \\
\Delta t
\end{bmatrix}, \ \ 
\HM{} = 
\begin{bmatrix}
1 & 0
\end{bmatrix}, 
\end{align*}
where the inputs to the system now consist of a discretized zero-order hold control acceleration $\uVec{\kCur}=2\cos(0.075\kCur)$, additive white process noise vector $\vVec{\kCur} \in \mathbb{R}^2$ with discrete time covariance $\Q{\kCur} \in \mathbb{R}^{2 \times 2}$, and additive white measurement noise $\wVec{\kCur}$ with discrete time covariance $\R{\kCur}$. 
Note that, given $\V$ and $\W$, the corresponding discrete time noise covariances are 
\begin{align}
\R{} &=  \frac{\W}{\Delta T}, \\
\Q{} &=  \int_{0}^{\Delta t} \int_{0}^{\Delta t} e^{\mathbf{A}\Delta t} \mathbf{\Gamma} \V \mathbf{\Gamma}^T e^{\mathbf{A}^T\Delta t} \delta(\tau_1 -\tau_2) d \tau_1 d\tau_2 \label{eq:VanLoans}
\end{align}
where the matrix expression for $\Q{}$ can be computed from $\mathbf{A}$, $\mathbf{\Gamma}$, $\V$, and $\delta T$ using Van Loan's method \cite{BrownHwang2012}. If $\A$ and $\Gamma$ are both known, then this relationship also allows us to design a full $2 \times 2$ positive definite symmetric covariance matrix $\Q{}$ by tuning the corresponding scalar continuous time process noise acceleration intensity $\V$ only.

We examine \BO{}-based Kalman filter tuning for the following cases:
\begin{enumerate}
\item tuning of unknown $\Q{}$ (i.e. unknown $\V$) with correctly known $\R{}$ and known model dynamics; 
\item simultaneous tuning of unknown $\Q{}$ and unknown $\R{}$, with correctly known model dynamics;
\end{enumerate}
\noindent All results here were obtained using the open source BayesOpt library
\cite{bayesopt} to apply the update steps and evaluate the surrogate and
acquisition functions, with the rest of the code developed by the authors in
C++.

\subsection{Case 1: Unknown $\Q{}$}

In this case, $\bovar = \V{}$ and the true process noise intensity for ground
truth simulations is $\V{} = 1$ (m/s$^2$)$^2$/s, which results in a true
discrete time process noise covariance of \begin{align*}
\Q{} = \begin{bmatrix}
 3 \times 10^{-4} & 5 \times 10^{-3} \\
 5 \times 10^{-3} & 0.1 
\end{bmatrix}.
\end{align*}
\BO{} was used to tune the Kalman filter design by searching over $\V{}$ and
using (\ref{eq:VanLoans}) to construct $\Q{}$, using true measurement noise
variance $\R{} = 1$ m$^2$ and the true dynamics model.

\begin{figure*} 
    \centering
  \subfloat[iteration 0 (seed data)]{%
       \includegraphics[width=0.31\textwidth]{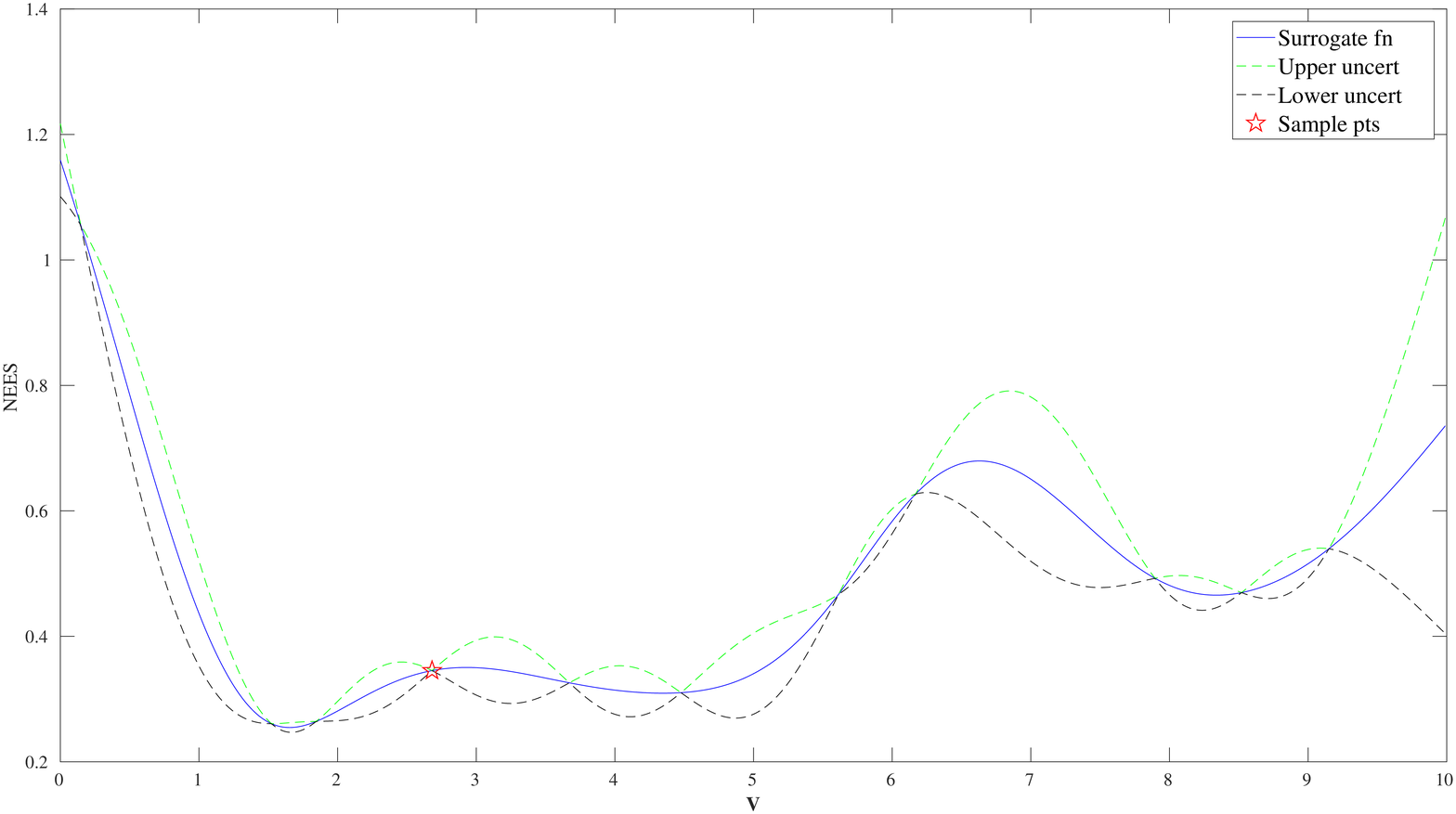}}
    \label{1a} 
  \subfloat[iteration 5]{%
    \includegraphics[width=0.31\textwidth]{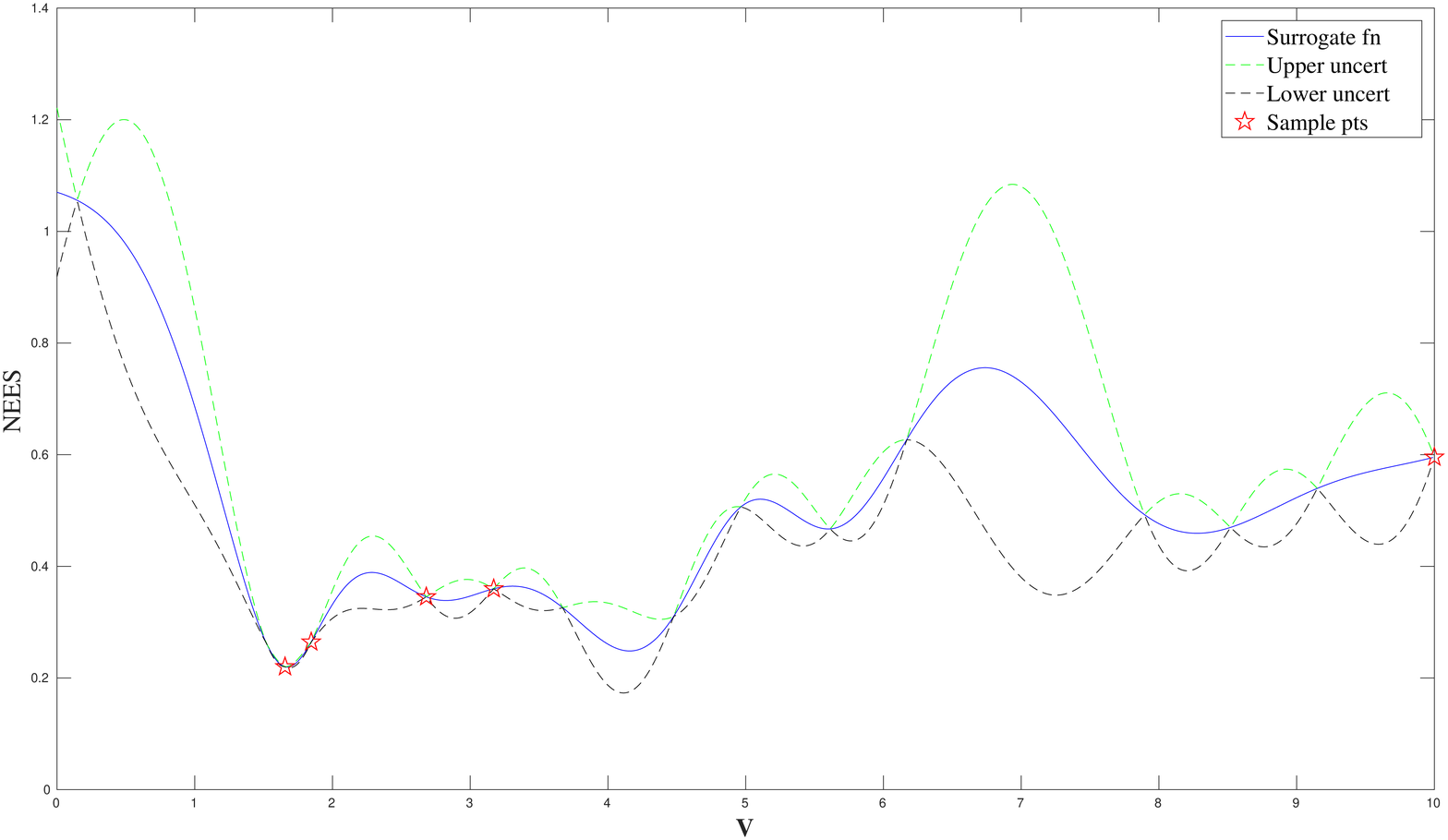}}
    \label{1b}
  \subfloat[iteration 10]{%
    \includegraphics[width=0.31\textwidth]{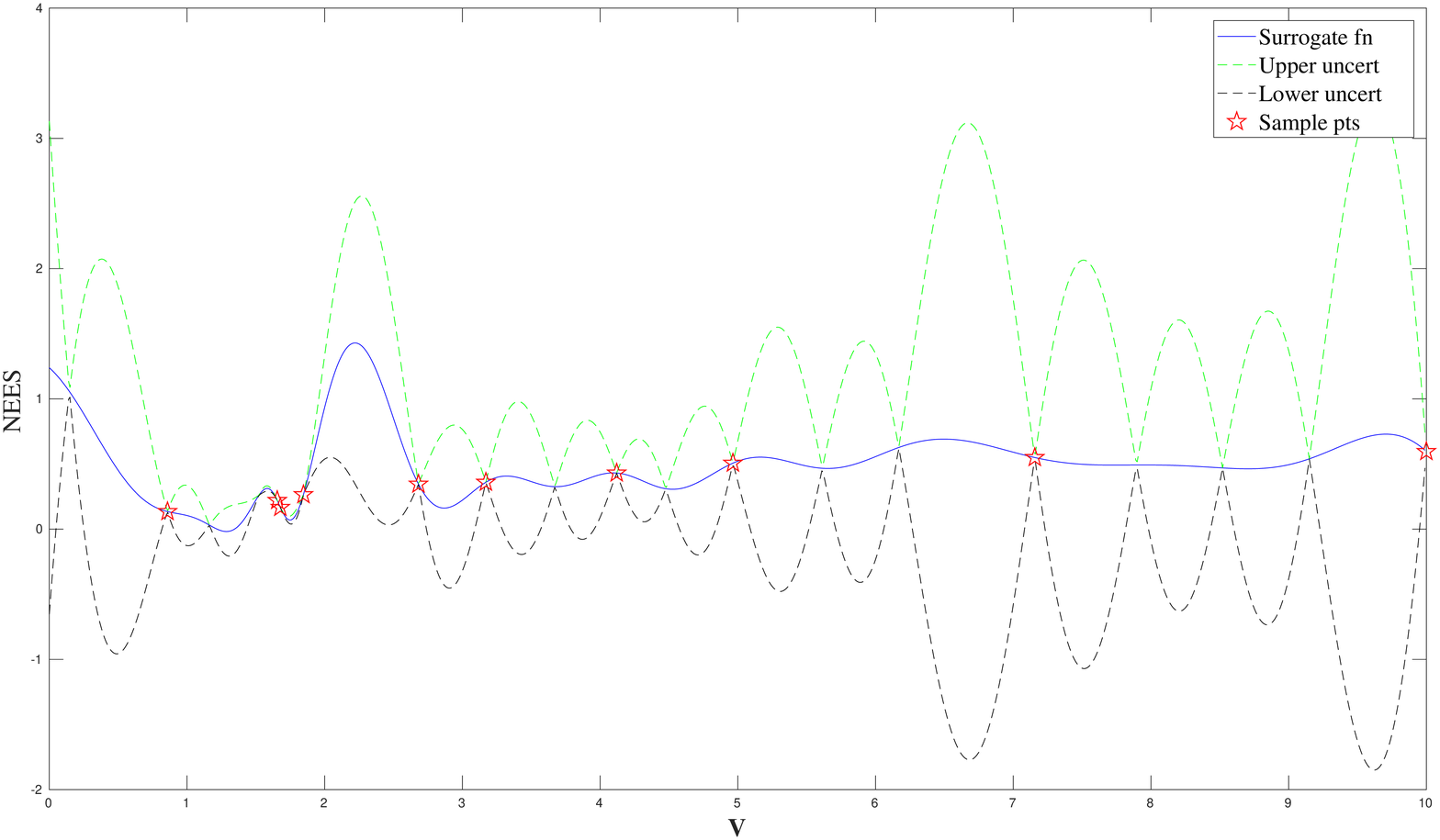}}
    \label{1c}  \\ 
  \subfloat[iteration 15]{%
    \includegraphics[width=0.31\textwidth]{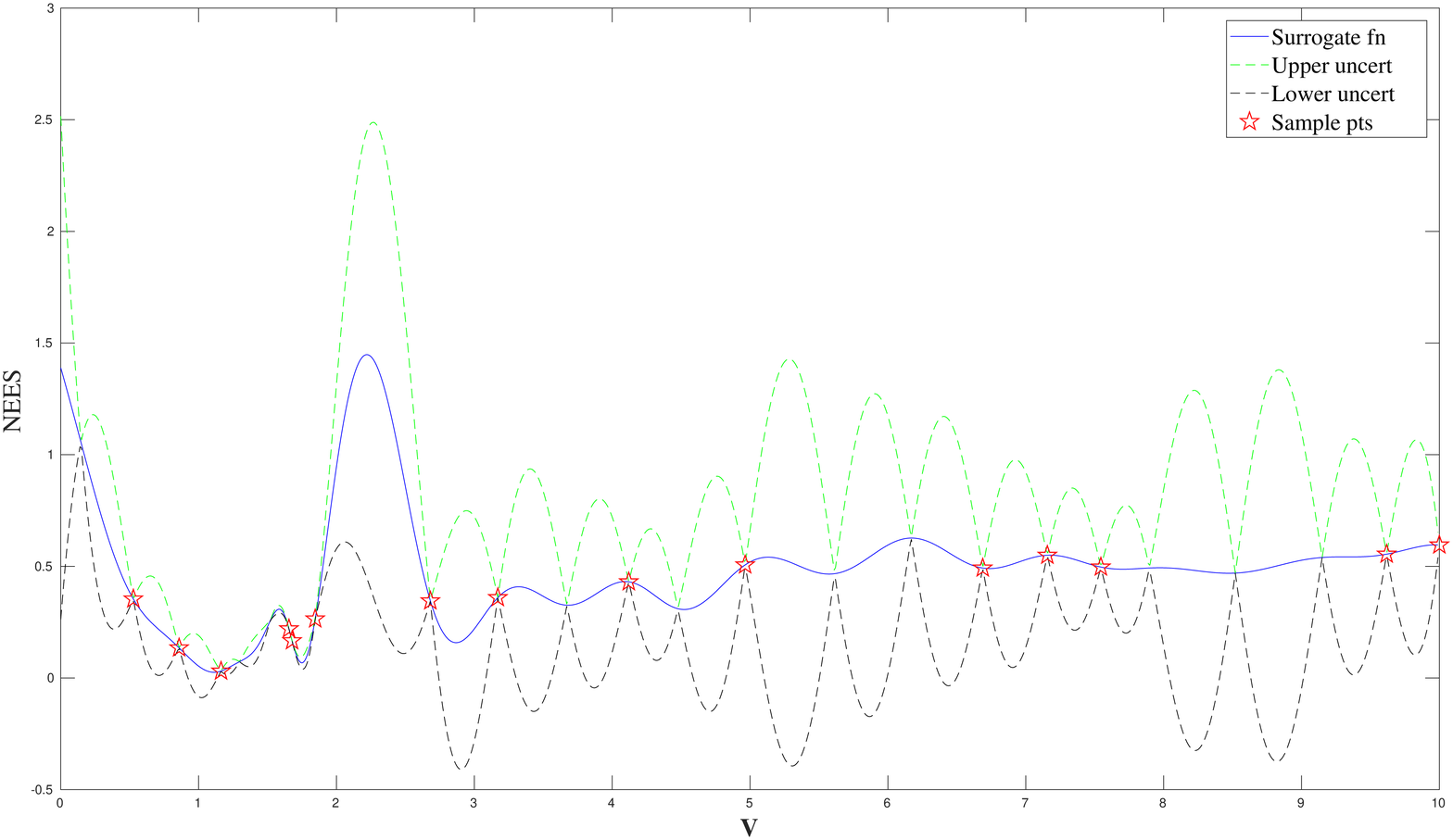}}
     \label{1d} 
     \subfloat[iteration 25]{%
       \includegraphics[width=0.31\textwidth]{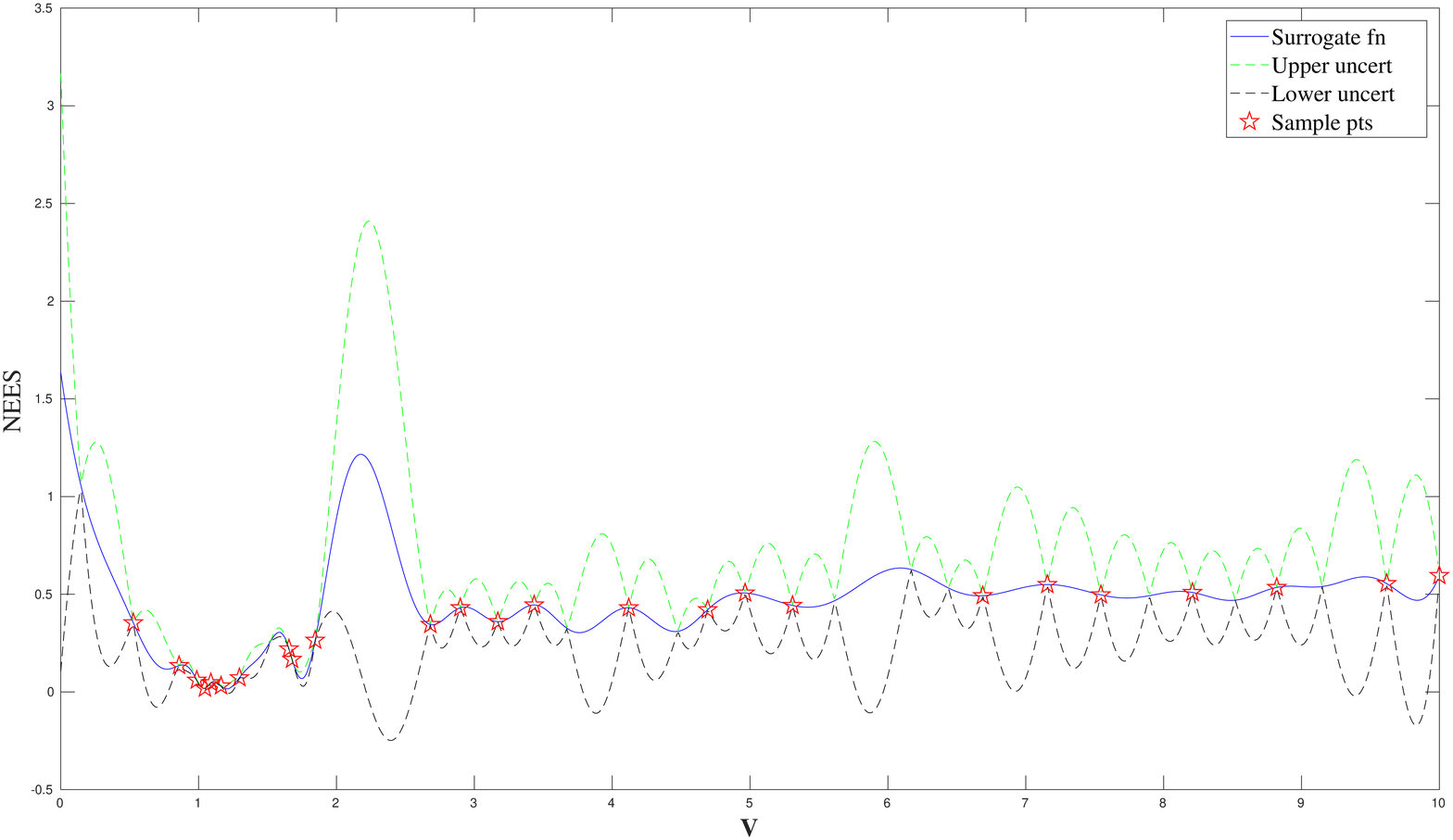}}
     \label{1e} 
     \subfloat[iteration 35]{%
       \includegraphics[width=0.31\textwidth]{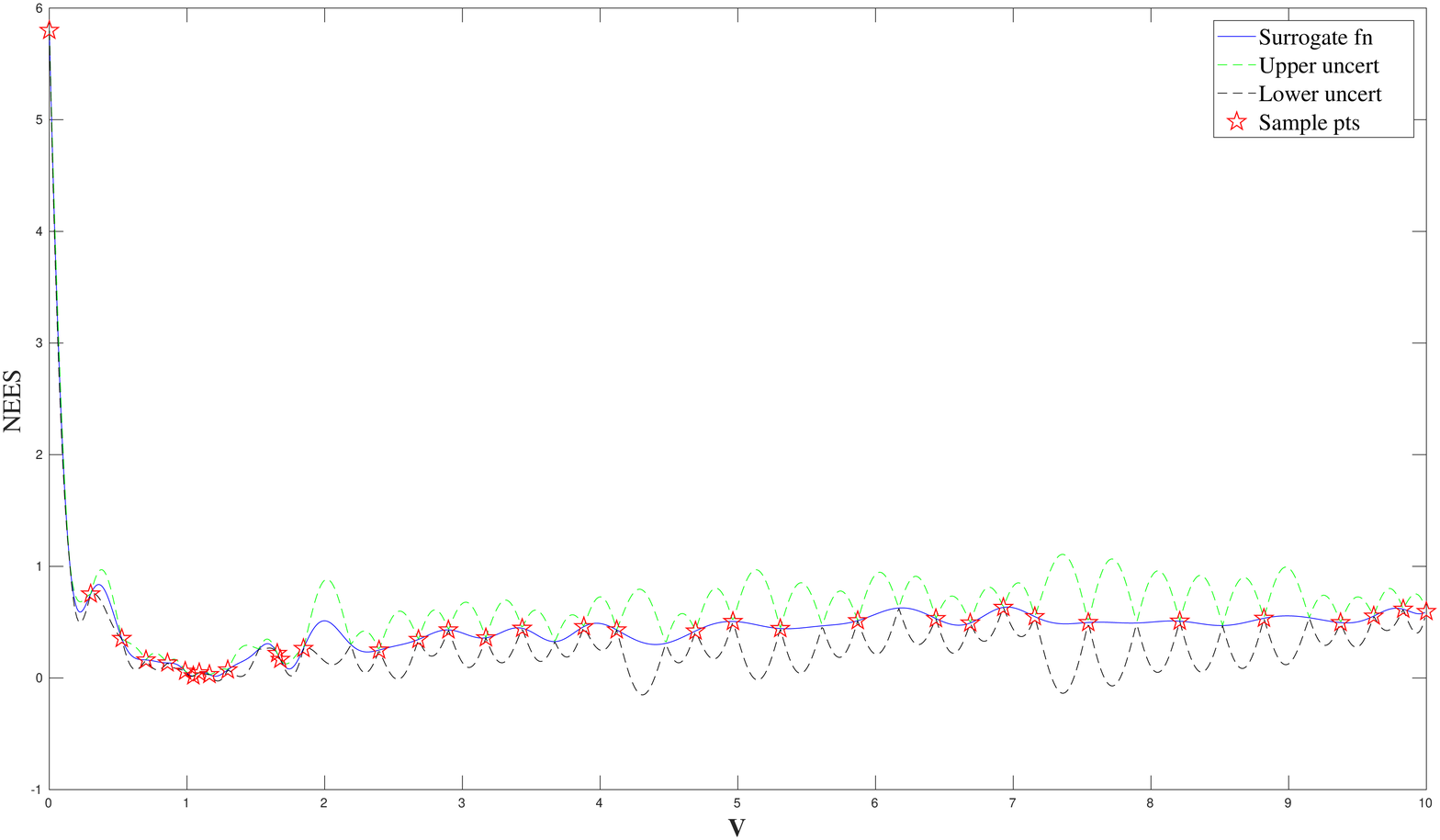}}
     \label{1e} 
  \caption{(a)-(f) \BO{} iterations for Case 1, showing surrogate GP model
  (top, with sampled points, mean and 2$\sigma$ bounds) and acquisition
  function (bottom).} \label{fig:case1nees} 
\end{figure*}

Figure \ref{fig:case1nees} shows six different iterations of a NEES-based \BO{}
search using the $\Jnees$ cost function over the range $\V{} = [0,10]$. In
these figures, $N=10$ Monte Carlo truth model simulations are used per $\Jnees$
evaluation, with $T=200$ time steps. 

The final result of this trial demonstrates that the minimum of the surrogate function
obtained through Bayesian optimization is very close to the ground truth, i.e.\ that
$\textbf{V} = 1$. Of particular value is also the uncertainty
bounds on the surrogate function that clearly demonstrate the uncertainty on these
parameters. The figures also demonstrate a clustering of sampling around the
true minimum of the objective function, but also a spread of samples in other places
to lower uncertainty of minima being in those regions.

\subsection{Case 2: Unknown $\Q{}$ and $\R{}$}
In this case, \BO{} was used to tune the Kalman filter design by searching over
$\V{}, \Q{}$ and using (\ref{eq:VanLoans}) to construct $\Q{}$, with
the noise variance $\R{}$ also to be estimated (cf. Case 1), while using
the true dynamics model for the robot.

\begin{figure*}
    \centering
  \subfloat[NEES results]{%
       \includegraphics[width=0.48\textwidth]{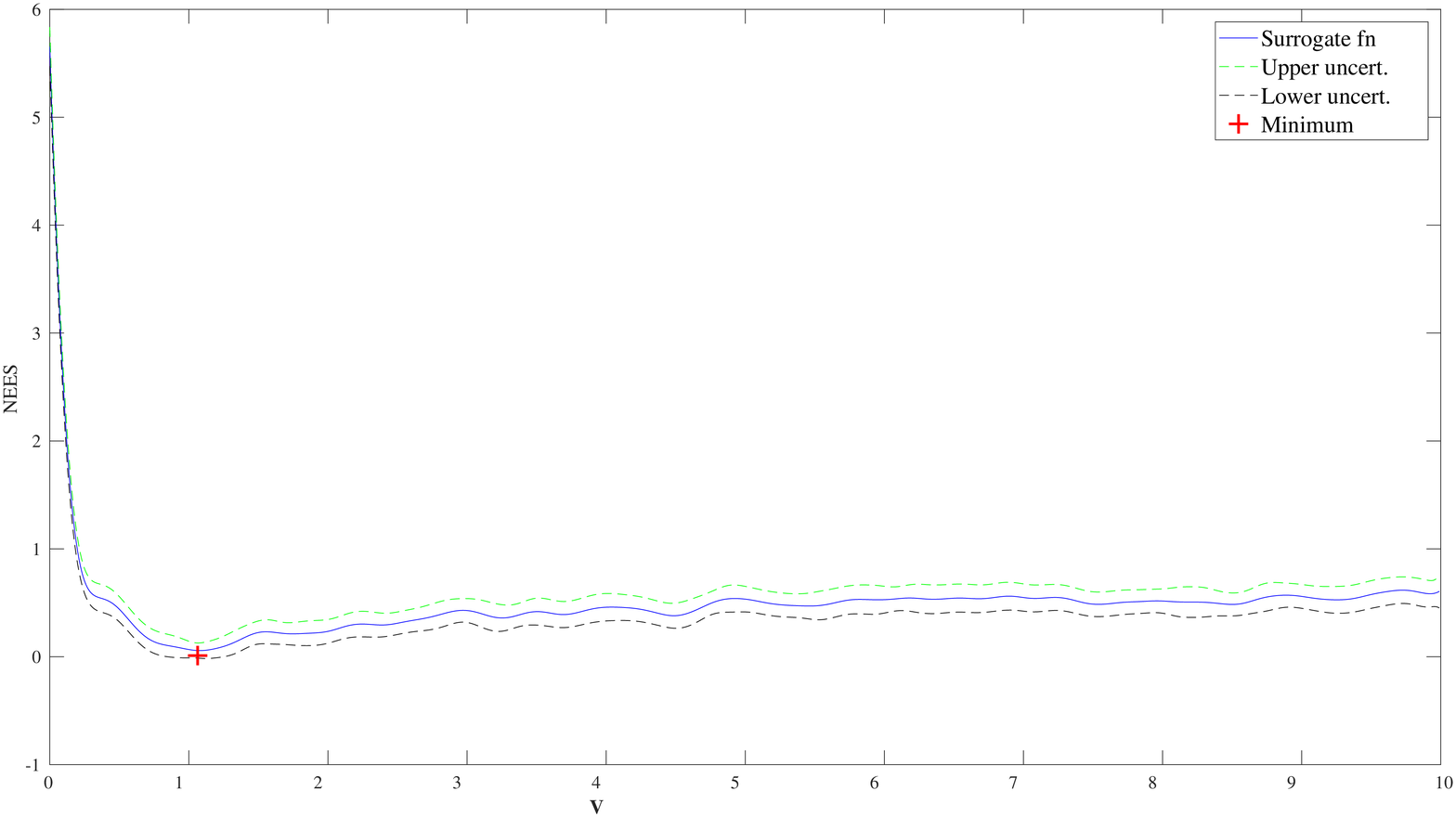}}
    \label{2a} 
  \subfloat[NIS results]{%
        \includegraphics[width=0.48\textwidth]{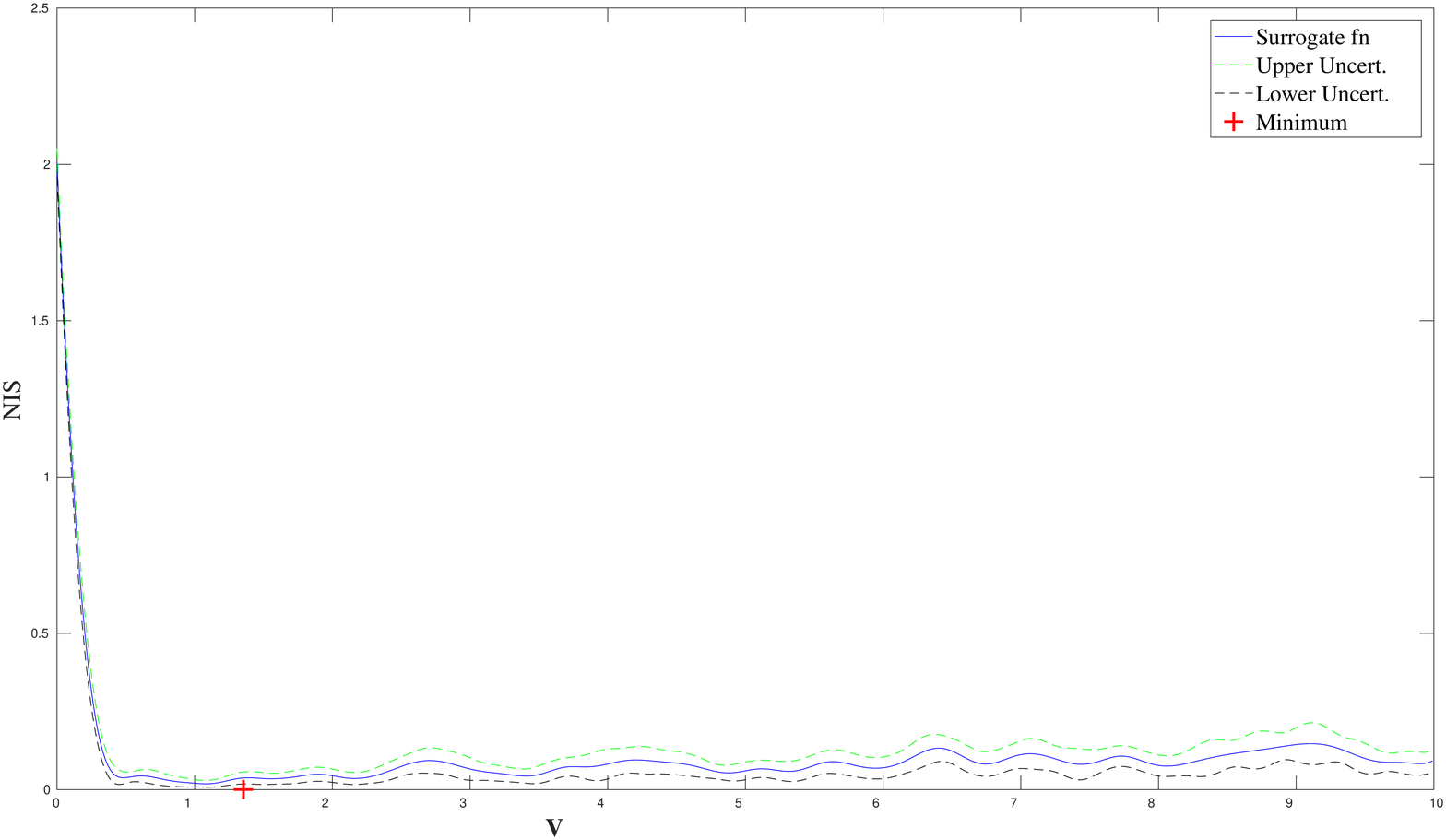}}
    \label{2b}
  \caption{Case 2, showing surrogate GP model with sampled points, mean
  and 2$\sigma$ bounds after 100 iterations of \BO{}: (a) NEES results; 
  (b) NIS results.} \label{fig:case2Q}
  \vspace{-0.5cm}
\end{figure*}

Figure \ref{fig:case2Q} shows \BO{}
search using the $\Jnees$ cost function over the range $\V{} = [0,10]$ in a 1-dimensional
cross section. In these figures, $N=10$ Monte Carlo truth model simulations are
used per $\Jnees$ evaluation, with $T=200$ time steps, just as in Case 1.

\begin{figure*}
    \centering
  \subfloat[NEES results]{%
       \includegraphics[width=0.45\textwidth]{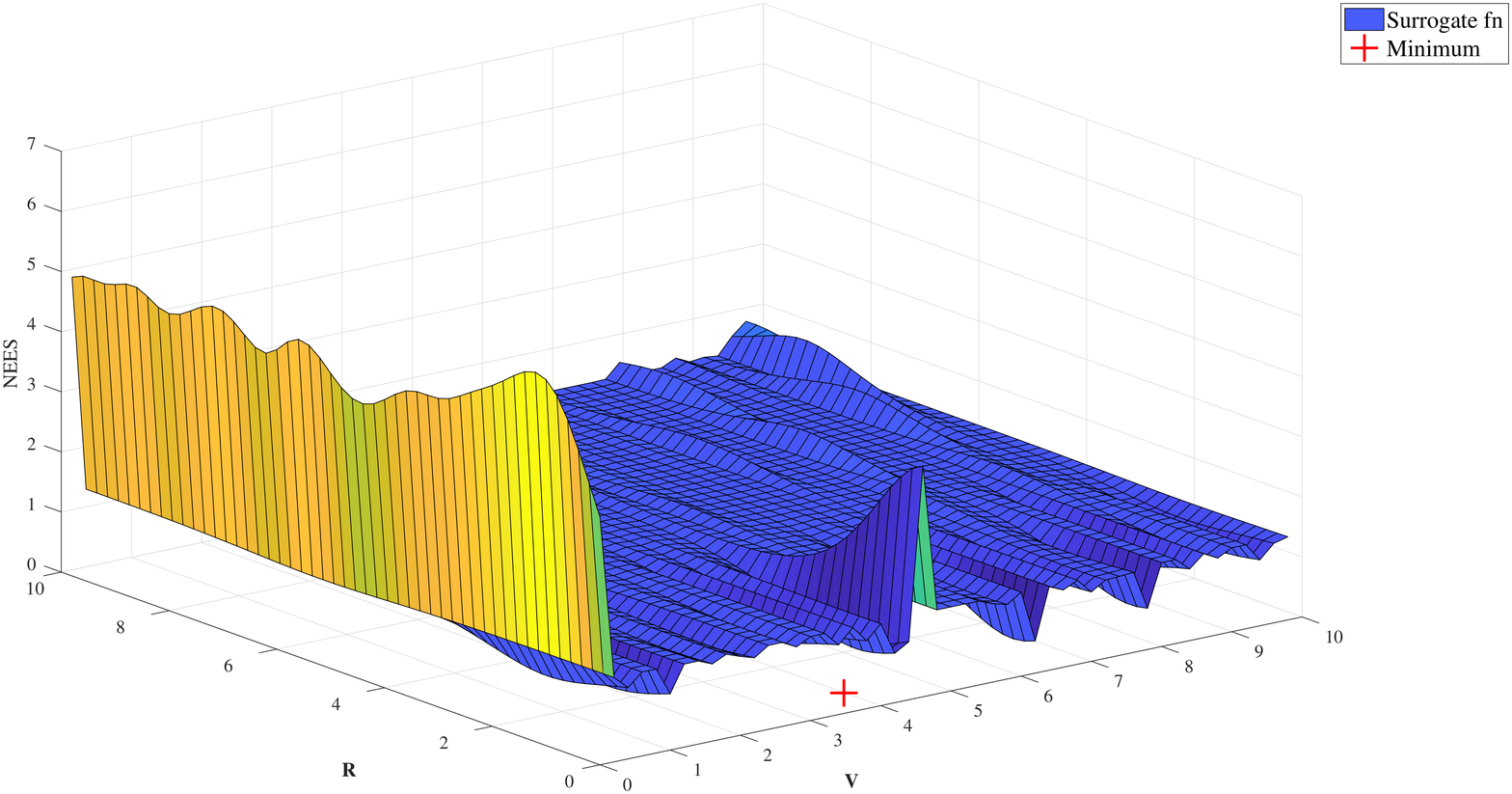}}
    \label{3a} 
  \subfloat[NIS results]{%
        \includegraphics[width=0.45\textwidth]{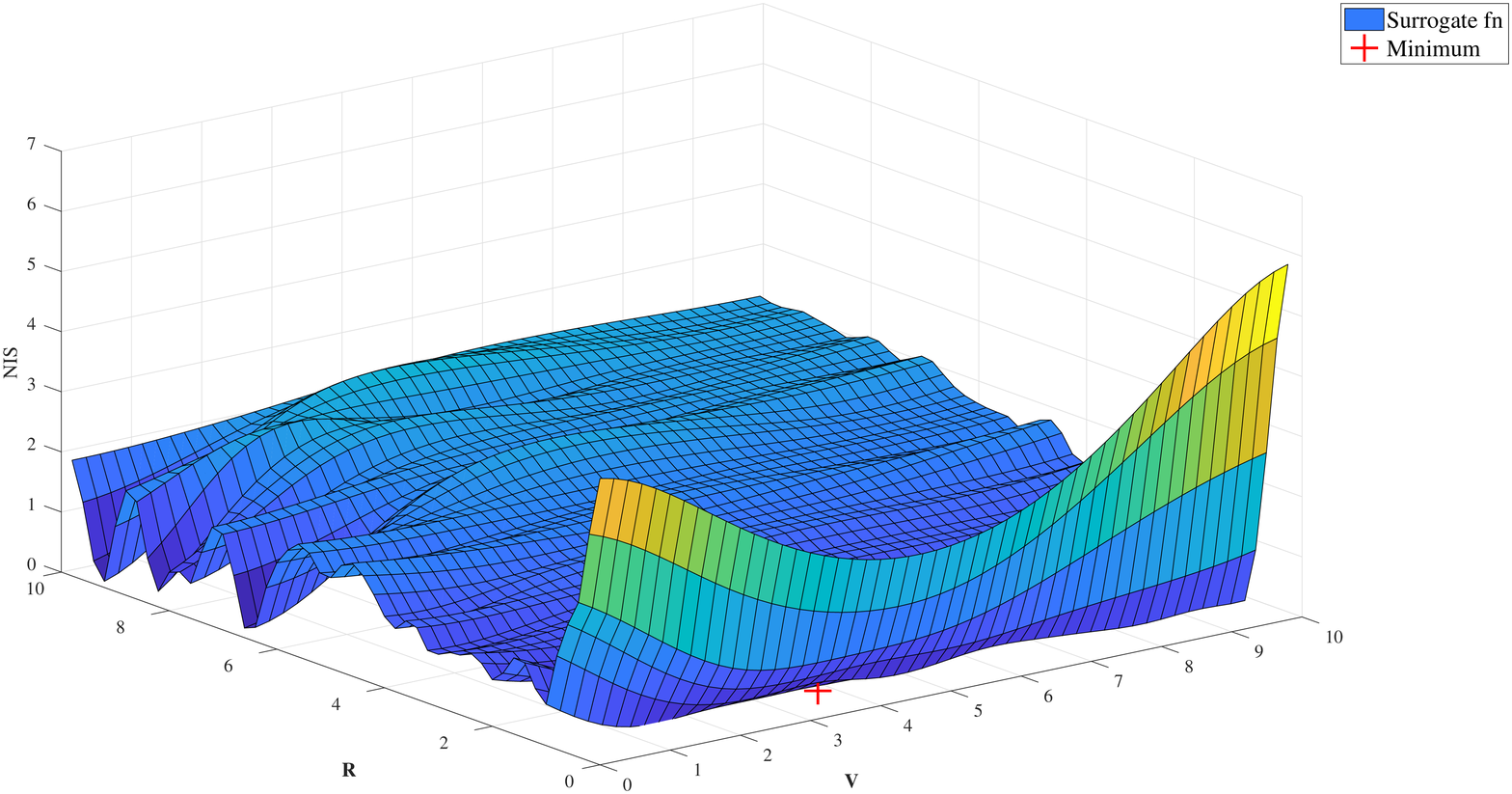}}
    \label{3b}
  \caption{Case 2, showing surrogate GP model with sampled points, mean and
  2$\sigma$ bounds after 100 iterations of \BO{}, for both \textbf{V} and
  \textbf{R} values. (a) shows the NEES results, while (b) shows the NIS
  results.} \label{fig:case2QR}
\end{figure*}

\begin{figure*}
    \centering
  \subfloat[$\mathbf{R} = 4.623, \mathbf{V} = 0.001$]{%
       \includegraphics[width=0.45\textwidth]{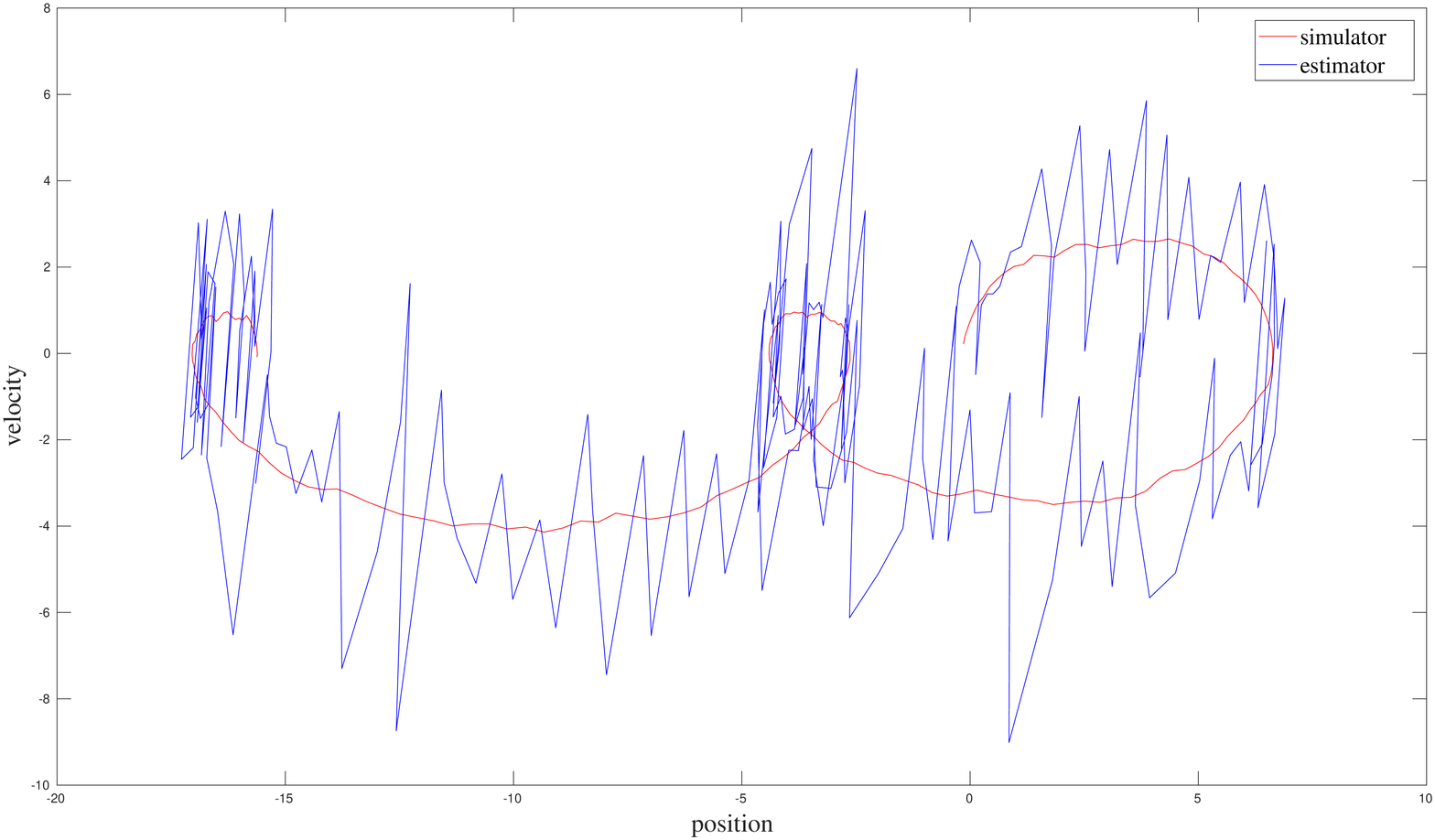}}
    \label{4a} 
  \subfloat[$\mathbf{R} = 3.925, \mathbf{V} = 0.587$]{%
        \includegraphics[width=0.45\textwidth]{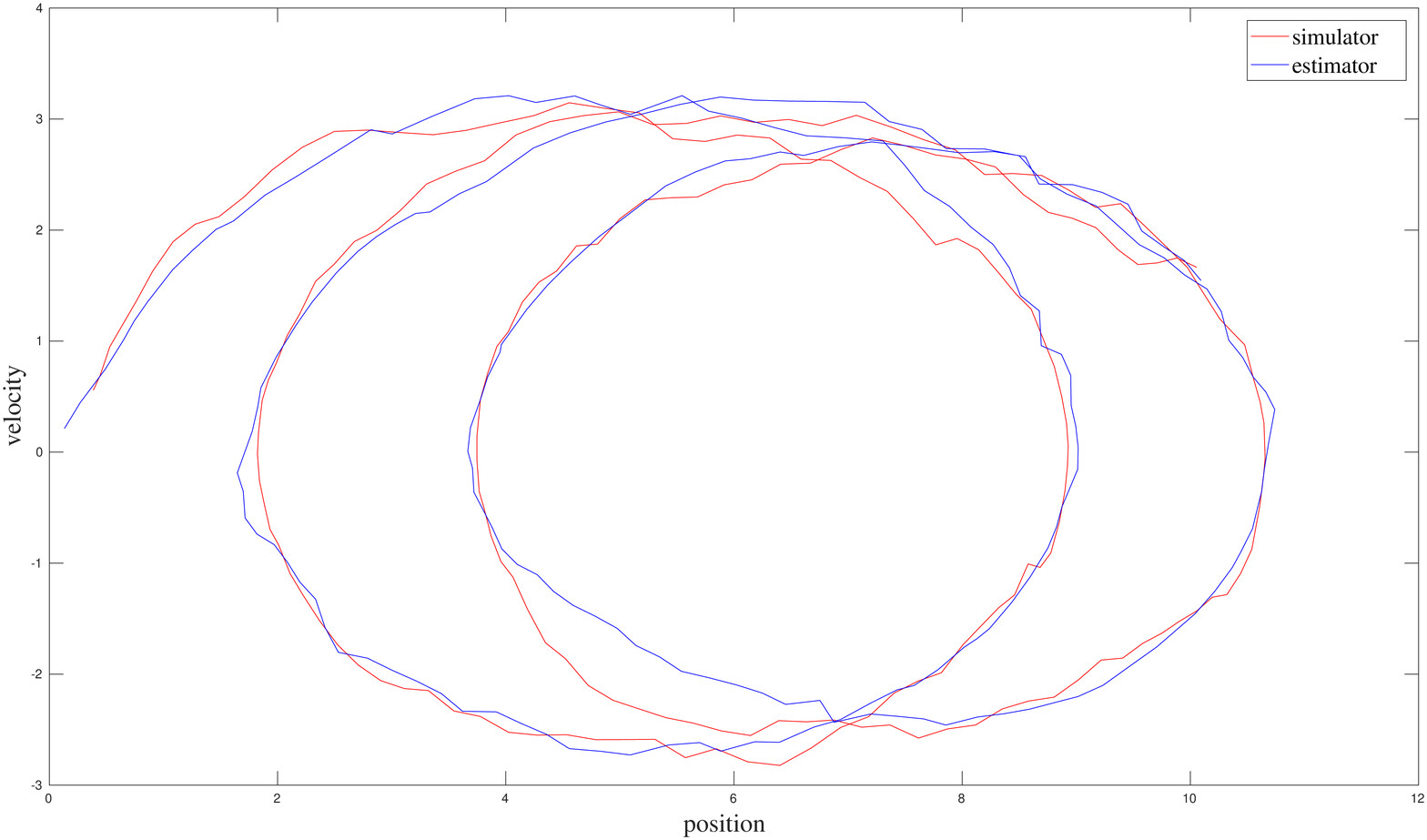}}
    \label{4b}
  \subfloat[$\mathbf{R} = 1.063, \mathbf{V} = 1.000$]{%
        \includegraphics[width=0.45\textwidth]{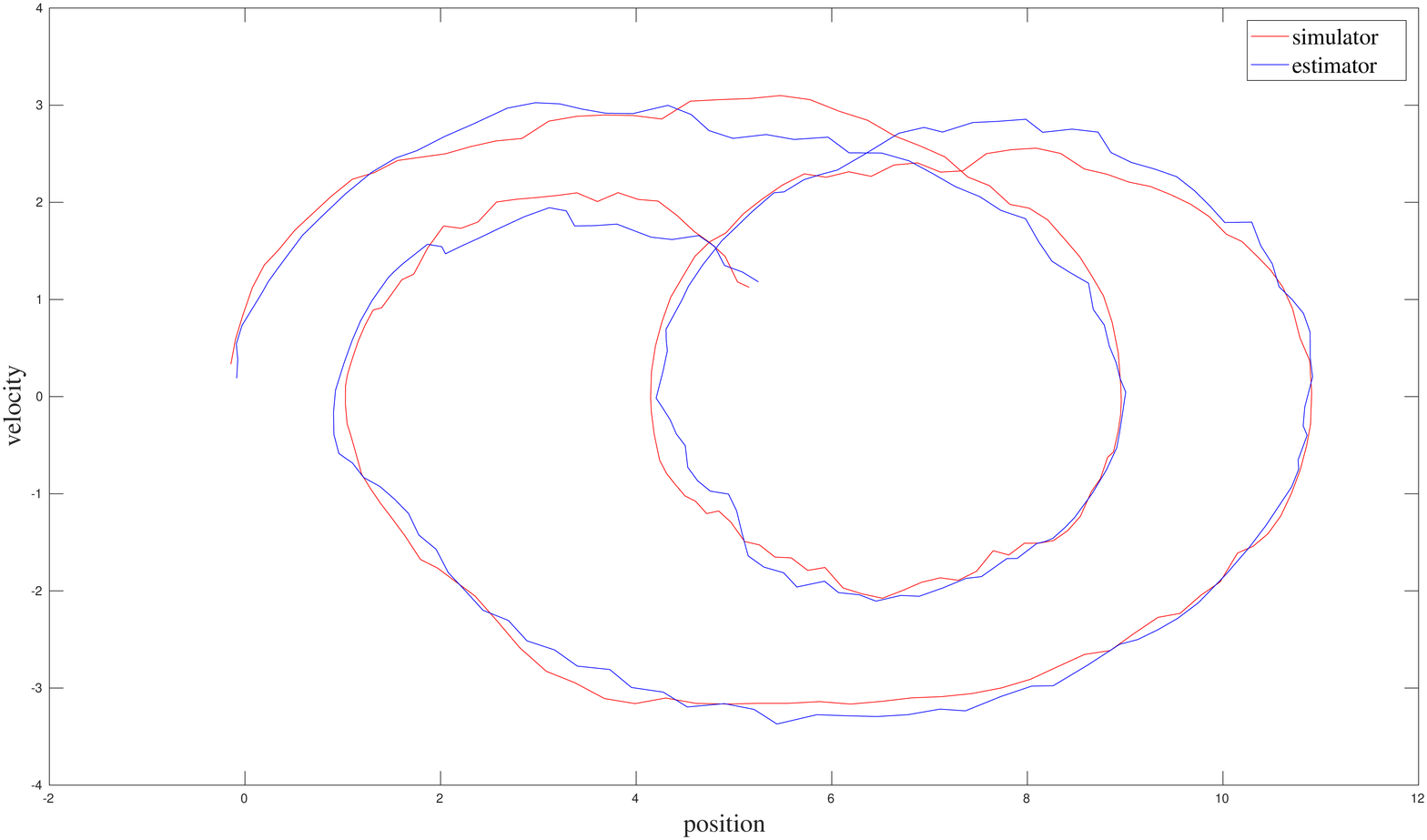}}
    \label{4c}
  \caption{Case 2 $\xi$ vs. $\dot{\xi}$ plots of the system dynamics overlaid
  with the estimated path from the Kalman filter.} \label{fig:case2states}
\end{figure*}

Figure \ref{fig:case2QR} shows the full 2-dimensional surrogate function after
100 iterations of Bayesian optimisation. We note that the results for
\textbf{R} appear more accurate than those for \textbf{V}, which exhibits a
large disturbance in the surrogate function near $\mathbf{V} \approx 4.5$,
leading to a large local minimum nearby. Inspection of the uncertainty around
the local minimum demonstrate that the surrogate function requires more
iterations to hone in on the global minimum. Fig. \ref{fig:case2states} shows
estimates of the path overlaid on the ground truth path in state space, demonstrating
that the two local minima are surprisingly close to one another
and underscoring the need for a global optimizer that outputs information on the various optima present in parameter space.

\section{Conclusions and Future Work} \label{sct:conc}
In this work we have developed a new approach to tuning Kalman filters
which lead to optimal estimates on the filter parameters, and have demonstrated
the method's success on the case of a single-dof robot in simulation. We have
shown that the uncertainty estimates resulting from the use of this method are
both a valuable addition to the classical optimization pipeline but also an
important point of consideration for determining the dependability of its
results.

In the future the authors will extend this work by turning to other
estimation problems, including sensor calibration parameters for e.g.\ visual
simultaneous localization and mapping. This will require the extension of
the method to more complex linear and non-linear system models, as well as
demonstrating its effectiveness with real hardware and experimental data.
There is also a wealth of experimentation to be conducted in the study of
other cost functions, acquisition functions, kernels, and parameterizations.
Furthermore, so-called ``pre-whitening'' filters might be leveraged to possibly
speed the convergence of \BO{} to accommodate non-white noise processes, and other optimization methods besides DIRECT
might improve the estimate of the global optimum once the method
has reached a certain threshold. 
%
Finally, \BO{} will be evaluated against alternative auto-tuning approaches, such as maximum likelihood estimation using expectation maximization \cite{Bishop2006}, online adaptive noise covariance estimation, \cite{Akhlaghi2017, Maybeck-AES-1981}, reinforcement learning \cite{goodall2007intelligent}, and simplex-based optimization \cite{powell2002automated}. 


\bibliographystyle{IEEEtran}
\bibliography{fusion2018_bib}

\end{document}